\documentclass[11pt]{article} 
\usepackage{rldmsubmit,palatino}
\usepackage{graphicx}
\usepackage{amsmath,amssymb}
\usepackage{listings}
\usepackage{algorithm}
\usepackage{algorithmic}
\usepackage{tikz}
\usepackage{subfigure}
\usetikzlibrary{arrows,positioning,shapes,calc}

\title{Combining  Deep Learning Architectures for Information Gain Estimation and Reinforcement Learning for Multiagent Field Exploration}

\author{
Emanuele Masiero, Giuseppe Vizzari, Dimitri Ognibene\\
University of Milan-Bicocca\\
Milan, Italy\\
\texttt{e.masiero@campus.unimib.it}\\
\texttt{dimitri.ognibene@unimib.it} \\
\And
Vito Trianni\\
ISTC-CNR \\
Rome, Italy \\
}

\begin{document}

\maketitle

\begin{abstract}

Precision agriculture increasingly demands autonomous systems capable of efficiently inspecting crops to detect issues such as pests, diseases, or fruit damage. This work addresses the problem as an active exploration task, proposing a strategy that combines information-theoretic reasoning with deep learning to guide agents through complex, partially observable environments like agricultural fields. The goal is to maximize relevant knowledge acquisition while minimizing resource consumption.

We model the environment as a discretized grid, where each cell may conceal targets observable only from specific points of view (POVs). 
To tackle this challenge, two predictive deep learning models were developed: a LSTM belief model responsible for estimating cell states, and IM (Informativity Metrics) model, which evaluates uncertainty reduction to select the best next point of view.

Three agent types were investigated: a non-learning IG agent that greedily selects actions based on entropy reduction, a Single-CNN DQN agent trained on localized belief and uncertainty inputs, and a Double-CNN DQN agent that integrates broader spatial context for long-range planning. Agents operate in a simulated $N \times N$ field, receiving $3 \times 3$ local observations and navigating using a constrained action space.
A critical design element is the inclusion of a POV visibility mask, which maintains the Markov property and prevents agents from revisiting ambiguous or redundant areas.

Results show that the untrained agent performs competitively—highlighting the power of information-theoretic heuristics—while the Double-CNN DQN achieves the best exploration efficiency and uncertainty reduction. Ablation studies confirm that removing the visibility mask leads to severe performance degradation, reinforcing the importance of structured, high-level input representations.

This framework generalizes to other domains involving data-efficient exploration and partial observability, including environmental sensing and industrial inspection. Future work will explore curriculum learning, transformer-based models for belief tracking, and collaborative multi-agent policies with shared buffers and rewards. These extensions aim to further improve generalization, scalability, and robustness of exploration in real-world deployments.

    \end{abstract}
\keywords{Information gain, Precision agriculture, Reinforcement learning}

\acknowledgements{We are deeply indebted to Google DeepMind and the Weinberg Institute for Cognitive Science for their generous support of RLDM2017. RTF template handmade, to the extent that we could emulate LaTeX. }  

\newpage
\section{Introduction}

Precision agriculture increasingly depends on autonomous systems to conduct timely and efficient crop inspection, particularly for tasks such as pest detection, disease identification, or fruit quality assessment. These scenarios demand not only accurate perception but also strategic sensing, where agents must autonomously explore large-scale environments while minimizing resource consumption, such as time, energy, and computational cost~\cite{zhang2021review, liu2020wholefield}.

In this context, we address the problem as an \emph{active exploration} task, where an agent operates within a discretized simulated environment representing an agricultural field. Each cell in the grid may contain targets (e.g., damaged plants) and can be inspected from nine predefined POVs. Observations from these fixed viewpoints provide rich but often redundant information, requiring agents to reason about uncertainty and information value to optimize exploration~\cite{singh2022multiagent}.

To guide exploration, we introduce a two-stage deep learning framework. A pre-trained recurrent neural network (LSTM) serves as a \emph{belief model} that incrementally updates a probabilistic representation of the environment based on sequential observations. This model outputs both a predicted target map and an associated entropy map, which together define the \emph{expected information gain (IG)} for each admissible observation~\cite{houthooft2016vime, sukhija2024maxinforl, case_2024}. These metrics form the foundation for adaptive exploration policies that prioritize high-uncertainty regions with potentially high informational value.

A central design choice involves incorporating a \emph{POV visibility mask} into the input representation. This addition enables the agent to preserve Markovianity even under partial observability by explicitly encoding which viewpoints have been previously visited. Without this information, agents tend to revisit the same regions and fail to construct consistent exploration policies. Embedding this visibility signal alongside belief and entropy maps significantly improves coverage and performance, particularly in early exploration stages.

Although our framework is tailored for crop monitoring, it generalizes to other domains requiring autonomous, data-efficient exploration, including robotics, environmental sensing, and industrial inspection~\cite{thrun2005probabilistic, burgard2005coordinated}. By combining information-theoretic exploration with deep reinforcement learning, this work offers a flexible and scalable foundation for future intelligent systems in complex, partially observed environments.

\section{Problem and Model Definition}

Autonomous exploration in large, partially observable environments is a central challenge in agricultural monitoring and related applications \cite{burgard2005coordinated, popovic2021informative}. In such scenarios, agents must efficiently gather information to reduce uncertainty about the environment while operating under strict resource constraints. One common approach is to guide agents toward regions expected to yield the highest information gain (IG) \cite{carbone2022monitoring, julian2014mutual}.

This work addresses a sequential decision-making problem where an agent explores a spatially structured grid environment to infer the presence of hidden targets (e.g., weeds or damaged crops) that are only visible from specific viewpoints. Each cell in the grid may contain one or more targets, and the agent receives noisy observations from multiple points of view (POVs). 
At each time step, the agent observes a local $3 \times 3$ neighborhood and can move in four directions. The objective is to minimize overall uncertainty—quantified via Shannon entropy—by selecting a sequence of actions that maximizes expected information gain across time.

\begin{equation}
\label{eq:entropy}
H(\boldsymbol{X}) = - \sum_i P(x_i) \log_2 P(x_i)
\end{equation}

We model the environment as a $N{\times}N$ grid where each cell may conceal targets observable from up to nine predefined POVs. The agent moves across the grid and receives partial observations from its current location. Its goal is to infer the number of targets in each cell with minimal exploration cost.

To guide decision-making, we use a pre-trained LSTM belief model that outputs a probability distribution over targets for each cell, updated with each new observation. 
The associated Shannon entropy quantifies uncertainty, and the agent selects actions maximizing expected information gain:
\begin{equation}
IG(a) = \sum_{\text{cells } c \in \mathcal{N}(a)} \left[ H(c) - H(c \mid \text{obs}_{t+1}(a)) \right]
\end{equation}
where \( \mathcal{N}(a) \) is the set of cells influenced by action \( a \), and \( \text{obs}_{t+1}(a) \) is the expected observation after taking action \( a \).

To guide exploration, we compare three agent architectures:
\begin{itemize}
    \item \textbf{Untrained IG Agent:} A non-learning baseline that selects the action with the highest computed information gain using the belief network.
    
    \item \textbf{Single-CNN DQN Agent:} A Deep Q-Network (DQN) that uses a CNN to process the local $3 \times 3$ observation window. The input includes belief states, entropy maps, and a binary mask indicating which POVs remain unobserved—crucial for maintaining the Markov property and avoiding perceptual aliasing.
    
    \item \textbf{Double-CNN DQN Agent:} An enhanced architecture with two CNN branches: one for the local $3 \times 3$ region and another for a wider context (e.g., $5 \times 5$ or larger). Feature maps are concatenated and passed to a fully connected layer to compute action-value estimates that incorporate both local and global spatial cues, supporting long-range planning and ambiguity resolution.
\end{itemize}

\section{Results}
We evaluated the agent's performance in a simulated environment, a $20 \times 20$ grid allowing, up to 1000 steps per episode.
All methods are compared against a baseline agent that moves randomly.

\begin{itemize}
  \item \textbf{Untrained IG Agent:}  
  This agent greedily selects actions that maximize instantaneous entropy reduction without any learning. Despite its simplicity, it achieves surprisingly effective exploration, maintaining strong performance across different environment sizes. Its efficiency highlights the power of information-theoretic guidance even without policy learning.

  \item \textbf{Single-CNN DQN Agent:}  
  This variant receives a local view of the belief map, its entropy, and a point-of-view (POV) mask. It performs comparably to the untrained IG agent in terms of exploration speed and coverage. However, when trained without the POV mask, the agent struggles—frequently revisiting previously observed areas—due to an inability to disambiguate similar local contexts. This confirms that the POV mask is critical for maintaining the Markov property in the agent's input.

  \item \textbf{Double-CNN DQN Agent:}  
  Equipped with a larger receptive field, this model integrates broader spatial context into its decision-making. It consistently outperforms the other agents, achieving more coherent trajectories, faster convergence, and greater uncertainty reduction. The extended spatial window allows it to balance local detail and global efficiency, albeit at a higher computational cost.

\end{itemize}

Additional experiments showed that agents trained directly on raw observations (the same input used by the belief network) perform decently but fall short of agents trained on abstract, uncertainty-aware inputs such as belief, entropy, and POV. This suggests that learning policies over structured, high-level representations yields more robust and efficient behaviors.

Overall, the results demonstrate that uncertainty-aware agents—especially those using the POV mask and broad spatial context—achieve more efficient, less redundant exploration and scale effectively to larger environments.

\begin{figure}[ht]
\centering
\subfigure[Agent with POV mask (average over 20 runs)]{
    \includegraphics[width=0.45\textwidth]{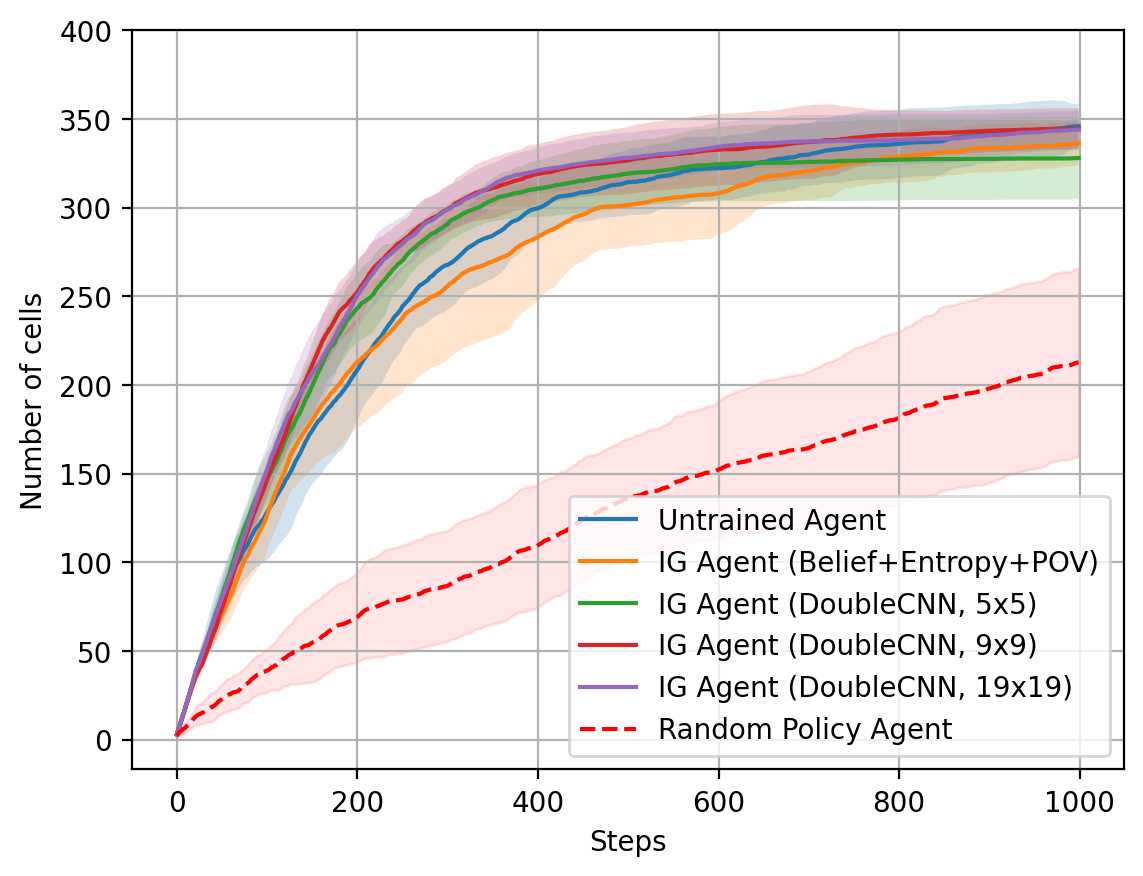}
    \label{fig:results_env20x20}
}
\hfill
\subfigure[Agent without POV mask (average over 20 runs)]{
    \includegraphics[width=0.45\textwidth]{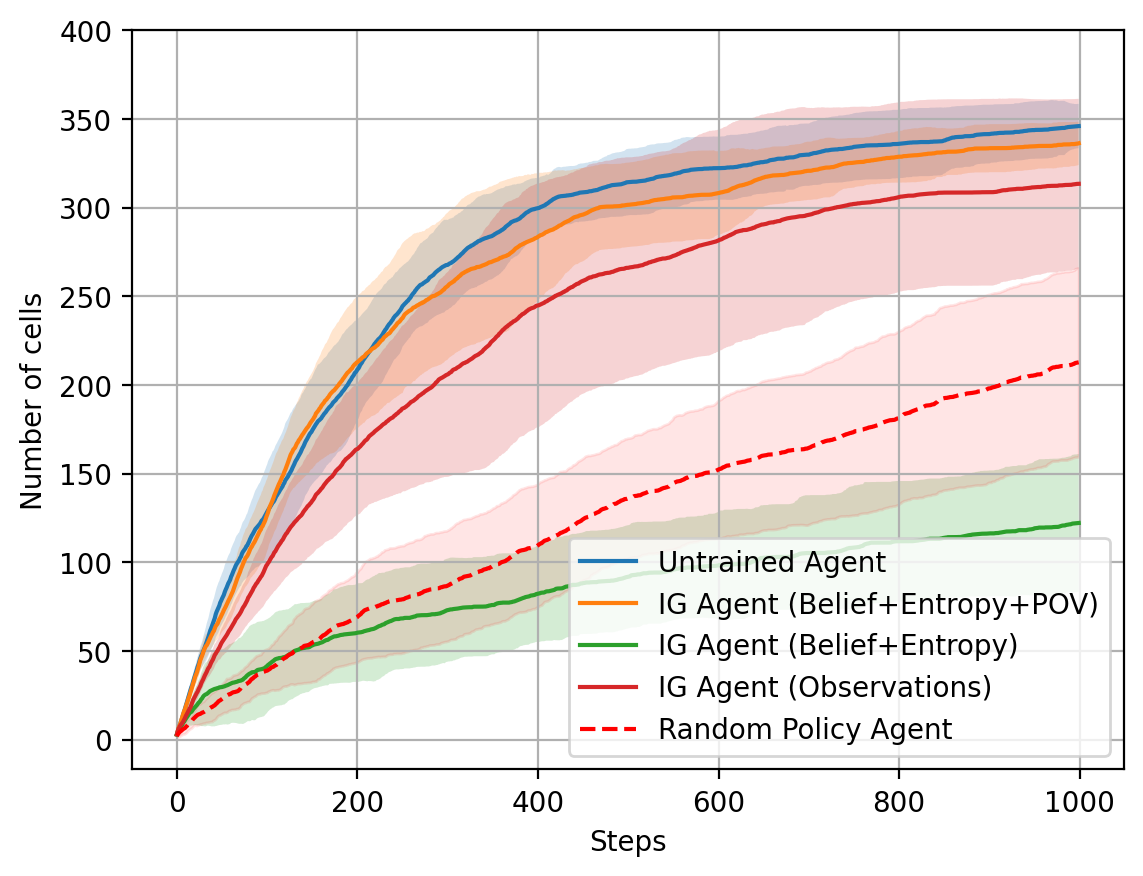}
    \label{fig:results_env20x20_with_and_without_pov}
}
\caption{Cells with correct target prediction at each step}
\end{figure}

\section{Conclusions}
This work addressed the challenge of active exploration in partially observable spatial environments, where an agent incrementally uncovers a hidden grid through partial observations. A pre-trained LSTM maintains a probabilistic belief over each cell, and the resulting entropy map guides the agent toward actions with high expected information gain. A simple non-learning agent that greedily maximizes instantaneous entropy reduction already performs surprisingly well, highlighting the effectiveness of information-theoretic strategies. Building on this, we trained a Deep Q-Network using a compact input composed of the belief, its entropy, and a POV mask. 

This representation was designed to be Markov-sufficient, capturing all the necessary information to predict future states without relying on full observation histories. Ablation studies demonstrated that removing the POV mask leads to policy failure, as the agent revisits visually ambiguous states and fails to progress. While a single-CNN variant matches the performance of the untrained heuristic, the best results were achieved with a Double-CNN architecture, which incorporates a broader spatial context and more effectively guides exploration—resembling map-based planning systems. Interestingly, the untrained heuristic agent scales nearly as well in larger environments, underscoring the general robustness of entropy-driven control. 

In summary, this work shows that combining probabilistic belief tracking with deep reinforcement learning and information-theoretic objectives leads to an efficient and scalable framework for exploration in complex environments. 

Future research will aim to initialize learning with informative-model priors to replace $\varepsilon$-greedy behavior, incorporate adaptive memory and intrinsic motivation mechanisms, enable safe multi-agent exploration through uncertainty-aware control, and design scalable latent state abstractions that preserve Markovian properties in large-scale POMDPs.

\bibliographystyle{plain}
\bibliography{rldm} 

\end{document}